\documentclass{article}
\usepackage{spconf,amsmath,graphicx,hyperref,xcolor,enumitem,booktabs,multirow,pifont,hyperref,tabularx}
\usepackage[table]{xcolor}
\newcommand{\cmark}{\ding{51}}
\newcommand{\xmark}{\textcolor{gray!50}{\ding{55}}}

\title{CLSC DETR: Reliable Candidate Ranking via Cross Layer Geometric Support for UAV Small Object Detection}
\name{Junyan Lin}
\address{South China Agricultural University\\
Guangzhou, China\\
\texttt{ljygarcia@stu.scau.edu.cn}
}
\begin{document}
%
\maketitle
\begin{abstract}
Unmanned aerial vehicle (UAV) object detection is critical for applications such as target search, where accurate detection of small objects in complex aerial scenes remains challenging.
The limited spatial extent, dense distribution, and frequent occlusion of small objects make reliable candidate ranking particularly difficult.
Existing Detection Transformer (DETR) based methods improve ranking by estimating localization quality from individual queries and incorporating it into classification scores.
However, a single query often lacks sufficient geometric evidence for small objects with weak boundary cues, resulting in unreliable quality estimation and unstable ranking.
To address this limitation, we propose Cross Layer Local Support and Consistency Calibration for DETR, termed CLSC DETR.
Specifically, the Cross Layer Local Support module establishes correspondences between final layer queries and intermediate layer candidates to aggregate complementary geometric evidence for more reliable localization quality estimation, while the Classification and Localization Consistency Calibration module adaptively adjusts classification scores according to localization quality and classification reliability to improve candidate ranking.
Experiments show that CLSC DETR improves AP and AP$_{75}$ over the baseline by 1.5\% and 2.0\% on VisDrone, respectively, while achieving consistent improvements on UAVDT.
\end{abstract}


\section{Introduction}
\label{sec:intro}

Recent advances in visual representation learning have driven progress across a broad range of vision tasks, including structured human generation and customizable virtual dressing~\cite{shen2024imagpose,shen2025imagdressing}. In aerial perception, unmanned aerial vehicles (UAVs) have been widely adopted owing to their flexible deployment and wide area observation capabilities, where reliable small object detection is essential for extracting critical target information~\cite{zhu2018visdrone}. However, objects in UAV imagery are often small, densely distributed, and heavily occluded, providing limited appearance and boundary cues for accurate localization. These characteristics not only increase localization difficulty but also make candidate ranking unreliable, since classification confidence may not faithfully reflect bounding box quality. Therefore, reliable localization quality estimation is critical for promoting well localized predictions and improving UAV small object detection.

\begin{figure}[t]
    \centering
    \includegraphics[width=0.97\linewidth]{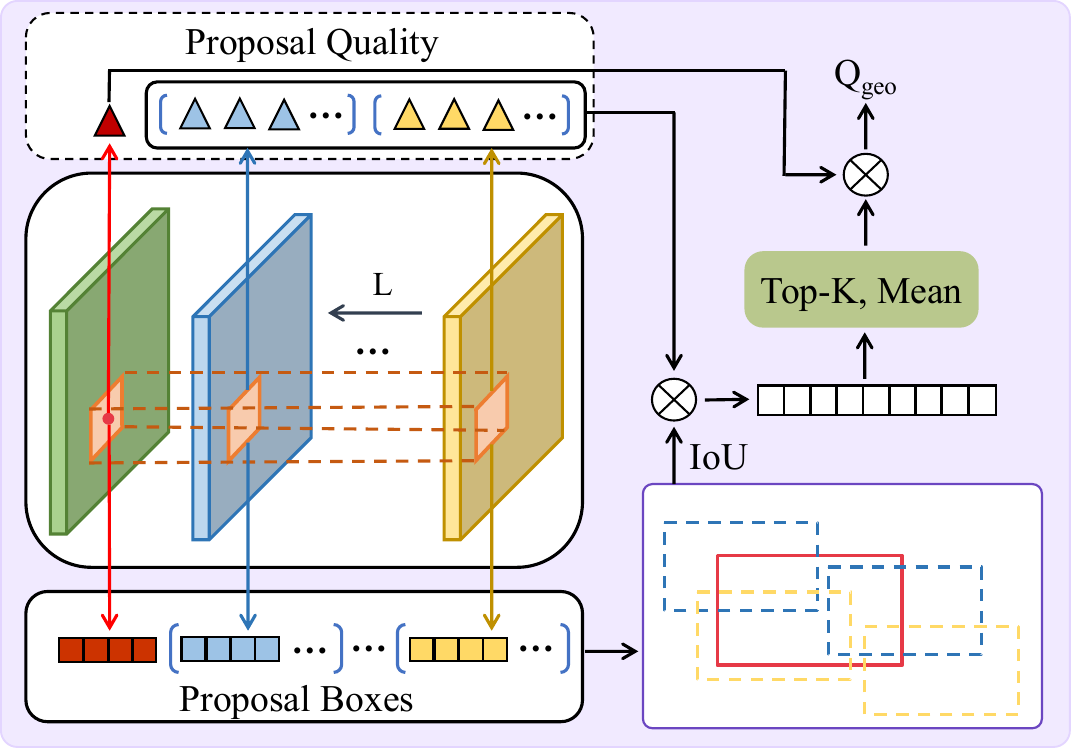}
    \vspace{-0.5cm}
    \caption{Structure of the Cross-Layer Local Support Module. $\otimes$ is the multiplication.}
    \label{fig:method1}
\vspace{-0.7cm}
\end{figure}

Existing studies have improved Detection Transformer (DETR)~\cite{carion2020detr} mainly through architectural design and feature representation. 
DINO enhances query learning through contrastive denoising training and mixed query selection~\cite{zhang2023dino}, while RT-DETR introduces an efficient hybrid encoder to balance detection accuracy and inference efficiency~\cite{zhao2024rtdetr}. MI-DETR further improves feature utilization through a parallel multi time inquiry mechanism~\cite{nan2025midetr}. For UAV imagery, UAV-DETR combines frequency enhancement, multiscale feature fusion, and semantic calibration to strengthen small object representations~\cite{zhang2025uavdetr}. HEDS-DETR and EFSI-DETR further preserve fine grained information through high frequency semantics, geometric priors, and frequency spatial interaction~\cite{peng2025hedsdetr,xia2026efsidetr}. Although these methods improve feature representation, candidate ranking still depends largely on classification scores, which are not necessarily consistent with the localization quality of the predicted boxes.

\begin{figure*}[t]
    \centering
    \includegraphics[width=0.9\linewidth]{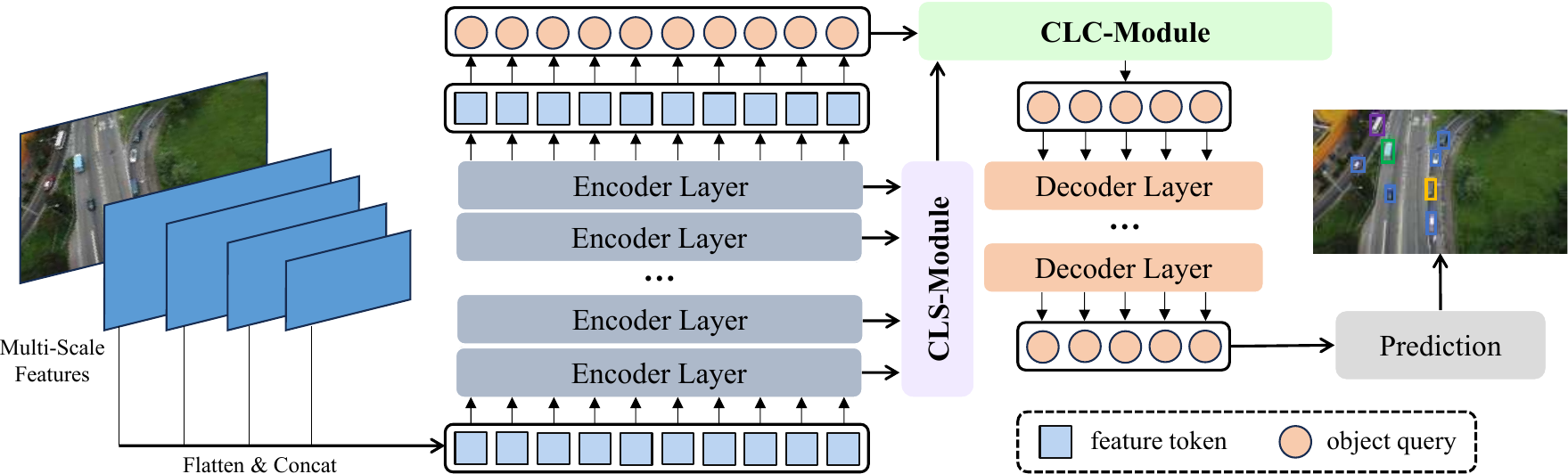}
    \vspace{-0.5cm}
    \caption{Overview of the \textbf{CLSC-DETR}. CLS-Module represents the Cross-Layer Local Support Module; CLC-Module denotes the Classification-Localization Consistency Calibration Module.}
    \label{fig:method_overview}
\vspace{-0.6cm}
\end{figure*}

Another line of research explicitly considers localization quality and classification localization consistency. IoU-Net predicts IoU as an estimate of bounding box localization quality~\cite{jiang2018iounet}, while Cascade-DETR predicts the expected IoU associated with each query and uses it to calibrate classification scores~\cite{ye2023cascadedetr}. Cal-DETR improves confidence calibration through uncertainty guided logit modulation and logit mixing~\cite{munir2023caldetr}. Rank-DETR introduces ranking oriented architectural designs, objectives, and matching costs to promote high quality predictions~\cite{pu2023rankdetr}, whereas Align-DETR improves classification localization consistency through task alignment and cross layer query alignment~\cite{cai2023aligndetr}. Despite these advances, localization quality is still estimated primarily from the representation of each individual candidate. For small objects, however, a single query contains limited and potentially noisy geometric information due to weak boundaries and insufficient spatial support. As a result, its localization quality estimate can fluctuate substantially, causing poorly localized predictions to receive high ranking scores while suppressing better localized candidates.

To address this limitation, we propose CLSC-DETR, a Cross Layer Local Support and Consistency Calibration framework for UAV small object detection. The key idea is to exploit complementary geometric evidence across network layers rather than relying solely on an individual query. Specifically, the Cross Layer Local Support Quality Estimation module establishes correspondences between final layer queries and intermediate layer candidates, and integrates their spatial overlap and localization quality to obtain more reliable geometric support. Based on the refined quality estimate, the Classification Localization Consistency Calibration module further adjusts the contribution of localization quality according to classification reliability, preventing uncertain geometric estimates from excessively suppressing semantically reliable candidates. Together, the two modules improve localization quality estimation and classification localization consistency, leading to more reliable candidate ranking.
Our main contributions are summarized as follows:
\begin{itemize}
    \item[\textcolor{black}{$\bullet$}] We propose a Cross Layer Local Support Quality Estimation module that aggregates complementary geometric evidence from intermediate candidates to improve localization quality estimation for small objects.
    
    \item[\textcolor{black}{$\bullet$}] We develop a Classification Localization Consistency Calibration module that adaptively balances classification confidence and localization quality according to classification reliability, enabling more reliable candidate ranking.
    
    \item[\textcolor{black}{$\bullet$}] Extensive experiments on VisDrone and UAVDT demonstrate the effectiveness of CLSC DETR. Compared with the baseline, our method improves AP and AP$_{75}$ by 1.5\% and 2.0\% on VisDrone, and by 3.0\% and 3.8\% on UAVDT, respectively.
\end{itemize}

\section{Proposed Method}
\label{sec:method}
\noindent\textbf{Overview.}
As shown in Fig.~\ref{fig:method_overview}, CLSC-DETR extends
MI-DETR~\cite{nan2025midetr} with Cross-Layer Local Support Quality Estimation and  Classification--Localization Consistency Calibration. The former leverages cross-layer support to improve the robustness of small-object query quality estimation, while the latter aligns classification scores with localization quality to enable more reliable candidate ranking.

\begin{table*}[t]
\centering
\caption{Comparison on the VisDrone-DET and UAVDT datasets.}
\label{tab:main_results}

\fontsize{9}{9.5}\selectfont
\setlength{\tabcolsep}{2.5pt}
\renewcommand{\arraystretch}{0.92}

\begin{tabularx}
  {\dimexpr\textwidth-2em\relax}
  {@{}lc*{10}{>{\centering\arraybackslash}X}@{}}

\toprule
& & \multicolumn{6}{c}{VisDrone-DET}
& \multicolumn{4}{c}{UAVDT} \\

\cmidrule(lr){3-8}
\cmidrule(lr){9-12}

Method
& Pub.Year
& AP
& AP$_{50}$
& AP$_{75}$
& AP$_{S}$
& AP$_{50}^{S}$
& AP$_{75}^{S}$
& AP
& AP$_{50}$
& AP$_{75}$
& AP$_{S}$ \\

\midrule

YOLOv10-X~\cite{wang2024yolov10}
& NIPS.2024
& 27.5
& 44.8
& 28.2
& 17.4
& 34.2
& 15.9
& 32.2
& 50.0
& 37.6
& 21.4 \\

YOLOv11-X~\cite{jocher2024yolo11}
& arXiv.2024
& 26.7
& 43.3
& 27.5
& 17.1
& 33.1
& 15.7
& 32.9
& 55.3
& 32.8
& 24.9 \\

\midrule

DINO~\cite{zhang2023dino}
& ICLR.2023
& 33.7
& 56.9
& 33.9
& 25.5
& 49.3
& 23.2
& 34.8
& 59.9
& 36.9
& 24.5 \\

Rank-DETR~\cite{pu2023rankdetr}
& NIPS.2023
& 32.7
& 54.5
& 33.0
& 24.4
& 46.5
& 22.2
& 33.1
& 58.2
& 35.4
& 23.5 \\

Align-DETR~\cite{cai2023aligndetr}
& BMVC.2024
& 32.9
& 55.3
& 32.8
& 24.9
& 48.1
& 22.2
& 34.8
& 57.1
& 39.9
& 22.8 \\

RT-DETR~\cite{zhao2024rtdetr}
& CVPR.2024
& 29.5
& 49.7
& 29.3
& 19.6
& 39.5
& 16.8
& 33.7
& 53.3
& 39.6
& 20.5 \\

MI-DETR~\cite{nan2025midetr}
& CVPR.2025
& 33.7
& 56.4
& 33.9
& 25.5
& 48.7
& 23.7
& 34.0
& 59.1
& 36.4
& 22.7 \\

\rowcolor{gray!15}
\textbf{Ours}
& --
& \textbf{35.2}
& \textbf{57.9}
& \textbf{35.9}
& \textbf{26.6}
& \textbf{49.9}
& \textbf{25.2}
& \textbf{37.0}
& \textbf{62.9}
& \textbf{40.2}
& \textbf{26.6} \\

\bottomrule
\end{tabularx}
\vspace{-0.4cm}
\end{table*}

\begin{table}[t]
\centering
\caption{Ablation experiments of the proposed modules.}
\label{tab:module_ablation}
\fontsize{9}{9.5}\selectfont
\setlength{\tabcolsep}{3.5pt}
\renewcommand{\arraystretch}{0.95}
\begin{tabularx}{\dimexpr\columnwidth-3em\relax}{@{}c*{5}{>{\centering\arraybackslash}X}@{}}
\toprule
\multirow{2}{*}{No.} & \multicolumn{2}{c}{Settings} & \multicolumn{3}{c}{Metrics} \\
\cmidrule(lr){2-3}\cmidrule(lr){4-6}
 & CLS & CLC & $AP$ & $AP_{50}$ & $AP_{75}$ \\
\midrule
1 & \xmark & \xmark & 33.7 & 56.3 & 33.9 \\
2 & \cmark & \xmark & 34.7 & 57.2 & 35.5 \\
3 & \xmark & \cmark & 34.6 & 57.7 & 35.0 \\
4 & \cmark & \cmark & 35.2 & 57.9 & 35.9 \\
\bottomrule
\end{tabularx}
\vspace{-0.2cm}
\end{table}

\subsection{Cross-Layer Local Support Quality Estimation}
In small-object scenarios, localization quality estimation based on individual queries is susceptible to feature noise. To obtain more stable estimates, we propose the Cross-Layer Local Support Quality Estimation module (CLS-Module), which enhances final-layer queries by aggregating geometric evidence from corresponding candidates across encoder layers.
As shown in Fig.~\ref{fig:method1}, a shared prediction head is applied to the intermediate encoder layers $\mathcal{M}$ and the final layer $L$ to ensure consistent predictions across layers. For each final-layer candidate $i$, we search a local neighborhood of radius $\rho$ around its corresponding position in each intermediate layer and collect the candidates into an auxiliary set $\Omega_i$.
For each auxiliary candidate $(\ell,j)\in\Omega_i$, where $\ell\in\mathcal{M}$ denotes the intermediate encoder layer index and $j$ denotes the candidate box index within that layer, its geometric evidence for final-layer candidate $i$ is defined as
\begin{equation}
g_{ij}^{\ell}
=
\operatorname{IoU}
\left(
\mathbf{b}_{i}^{L},
\mathbf{b}_{j}^{\ell}
\right)
q_{j}^{\ell},
\label{eq:local_geometric_evidence}
\end{equation}
where $\mathbf{b}$ and $q$ denote the predicted box and localization quality, respectively. Thus, only reliable auxiliary candidates that highly overlap with the final-layer candidate provide strong support.
To suppress noisy correspondences, we aggregate only the Top-$K$ candidates with the strongest geometric evidence:
\begin{equation}
\mathcal{T}_i
=
\operatorname{TopK}_{(\ell,j)\in\Omega_i}
\left(g_{ij}^{\ell}\right),
\qquad
G_i
=
\frac{1}{|\mathcal{T}_i|}
\sum_{(\ell,j)\in\mathcal{T}_i}
g_{ij}^{\ell},
\label{eq:topk_support}
\end{equation}
where $G_i$ denotes the aggregated cross-layer local support. The localization quality is then refined as
\begin{equation}
Q_i^{\mathrm{geo}}
=
\alpha G_i q_i^{L},
\label{eq:geometric_quality}
\end{equation}
where $\alpha$ is a scaling coefficient. A candidate obtains a high $Q_i^{\mathrm{geo}}$ only when it has reliable intrinsic localization and receives consistent cross-layer support, reducing quality fluctuations caused by single-query prediction errors.


\subsection{Classification--Localization Consistency Calibration}

We propose Classification--Localization Consistency Calibration
(CLC-Module), which adaptively adjusts the contribution of geometric
quality according to classification reliability.
For candidate $i$, let $\mathbf{p}_i=\{p_{i,c}\}_{c\in\mathcal{C}}$ and $\mathbf{z}_i$ denote its classification probability vector and logits, respectively. We estimate its classification reliability using the maximum classification score, classification margin, and entropy-based certainty:
\begin{equation}
\begin{aligned}
s_i &= \max_{c\in\mathcal{C}} p_{i,c}, \qquad
m_i = p_{i,c_i^{(1)}}-p_{i,c_i^{(2)}},\\
h_i &= 1-
\frac{\mathcal{H}(\operatorname{Softmax}(\mathbf{z}_i))}
{\log|\mathcal{C}|},\\
R_i^{\mathrm{cls}}
&=
\lambda_1\mathcal{N}(s_i)
+\lambda_2\mathcal{N}(m_i)
+\lambda_3\mathcal{N}(h_i),
\end{aligned}
\label{eq:classification_reliability}
\end{equation}
where $c_i^{(1)}$ and $c_i^{(2)}$ denote the categories with the highest and second-highest classification probabilities, respectively. $\mathcal{H}(\cdot)$ denotes Shannon entropy, $\mathcal{N}(\cdot)$ denotes image-wise normalization, and $\lambda_1$, $\lambda_2$, and $\lambda_3$ are the fusion weights of the three classification cues.
We then compare $R_i^{\mathrm{cls}}$ with the normalized geometric quality to compute the protection factor and final calibrated score:
\begin{equation}
\begin{aligned}
u_i
&=
\operatorname{ReLU}
\left(
R_i^{\mathrm{cls}}
-\mathcal{N}(Q_i^{\mathrm{geo}})
\right)
R_i^{\mathrm{cls}},\\
S_{i,c}
&=
p_{i,c}
\left(Q_i^{\mathrm{geo}}\right)^{
\beta(1-\gamma u_i)},
\end{aligned}
\label{eq:adaptive_calibration}
\end{equation}
where $u_i$ and $S_{i,c}$ are the protection factor and final category score, respectively; $\beta$ and $\gamma$ control the geometric contribution and protection strength. When classification is more reliable than geometric quality, $u_i$ reduces the geometric penalty to avoid suppressing valid candidates and improve ranking.

\section{Experiment and Analysis}
\label{sec:exp}
\noindent\textbf{Datasets.}
We evaluate CLSC-DETR on UAVDT~\cite{Du2018UAVDT} and VisDrone2019-DET~\cite{Zhu2022VisDrone}. For VisDrone, the official training and validation sets, containing 6,471 and 548 images, are used for training and evaluation, respectively. For UAVDT, we construct UAVDT-5.3K from its 30 training sequences. We use 24 sequences for training by sampling every four frames and the remaining six for validation by sampling every ten frames, yielding 4,793 and 503 images without sequence overlap. 

\noindent\textbf{Implementation Details.}
CLSC-DETR is built upon MI-DETR~\cite{nan2025midetr} with a ResNet-50 backbone~\cite{he2016resnet}. All models are trained for 90,000 iterations on two NVIDIA GeForce RTX 5090 GPUs with a batch size of 4. We use AdamW~\cite{loshchilov2019decoupled} with learning rates of $1\times10^{-4}$ and $1\times10^{-5}$ for the detector and backbone, respectively, and a weight decay of $1\times10^{-4}$. The learning rate is reduced to $1\times10^{-5}$ after 82,500 iterations, and gradients are clipped with a maximum norm of 0.1. We report the standard COCO metrics, including AP, AP$_{50}$, AP$_{75}$, and AP$_{S}$.
\subsection{Comparison with State-of-the-art Methods}
As shown in Table~\ref{tab:main_results}, CLSC-DETR achieves the best overall performance on both datasets. Compared with MI-DETR, it improves AP/AP$_{50}$/AP$_{75}$/AP$_{S}$ by 1.5\%/1.5\%/2.0\%/1.1\% on VisDrone and 3.0\%/3.8\%/3.8\%/3.9\% on UAVDT. These results validate the effectiveness of cross-layer geometric support and consistency calibration in improving localization quality estimation and candidate ranking. The larger gains on UAVDT may stem from its vehicle-dominated, concentrated category distribution and dense, scale-varying objects, which make candidate ranking more sensitive to localization quality.

\begin{table}[t]
\centering
\caption{Comparison of score calibration strategies.}
\label{tab:calibration_strategy}
\fontsize{9}{9.5}\selectfont
\setlength{\tabcolsep}{3.5pt}
\renewcommand{\arraystretch}{0.95}
\begin{tabularx}{\dimexpr\columnwidth-3em\relax}{@{}l*{3}{>{\centering\arraybackslash}X}@{}}
\toprule
Strategy & $AP$ & $AP_{50}$ & $AP_{75}$ \\
\midrule
Fixed exponent ($\eta=0.5$) & 34.8 & 57.2 & 35.4 \\
Linear fusion ($\omega=0.65$) & 34.8 & 57.3 & 35.4 \\
Ours & 35.2 & 57.9 & 35.9 \\
\bottomrule
\end{tabularx}
\vspace{-0.2cm}
\end{table}

\begin{table}[t]
\centering
\caption{Sensitivity analysis of key hyperparameters in CLSC-DETR. Only the specified parameter is varied in each group, while all other settings remain fixed.}
\label{tab:hyperparameter_sensitivity}
\fontsize{9}{9.5}\selectfont
\setlength{\tabcolsep}{3.5pt}
\renewcommand{\arraystretch}{0.95}
\begin{tabularx}{\dimexpr\columnwidth-3em\relax}
{@{}*{5}{>{\centering\arraybackslash}X}@{}}
\toprule
Parameter & Value & AP & AP$_{50}$ & AP$_{75}$ \\
\midrule
\multirow{4}{*}{$K$}
& 1 & 35.1 & 57.5 & \textbf{36.1} \\
& 2 & \textbf{35.2} & \textbf{57.9} & 35.9 \\
& 4 & 35.0 & 57.3 & 35.9 \\
& 8 & 34.8 & 57.0 & 35.6 \\
\midrule
\multirow{4}{*}{$\alpha$}
& 0.5 & 34.9 & 57.0 & 35.6 \\
& 1.0 & \textbf{35.2} & \textbf{57.9} & 35.9 \\
& 2.0 & 34.9 & 57.6 & \textbf{36.0} \\
& 3.0 & 34.8 & 57.7 & 35.5 \\
\midrule
\multirow{4}{*}{$\beta$}
& 0.10 & 35.0 & \textbf{58.1} & 35.6 \\
& 0.20 & \textbf{35.2} & 57.9 & 35.9 \\
& 0.30 & 35.1 & 57.5 & \textbf{36.0} \\
& 0.50 & 34.8 & 56.4 & 35.9 \\
\midrule
\multirow{4}{*}{$\gamma$}
& 0.8 & 35.1 & 57.5 & \textbf{36.0} \\
& 1.2 & \textbf{35.2} & \textbf{57.9} & 35.9 \\
& 4.0 & 34.8 & 57.3 & 35.5 \\
& 8.0 & 34.5 & 57.0 & 35.0 \\
\bottomrule
\end{tabularx}
\vspace{-0.2cm}
\end{table}

\begin{figure}[t]
    \centering
    \includegraphics[width=0.95\linewidth]{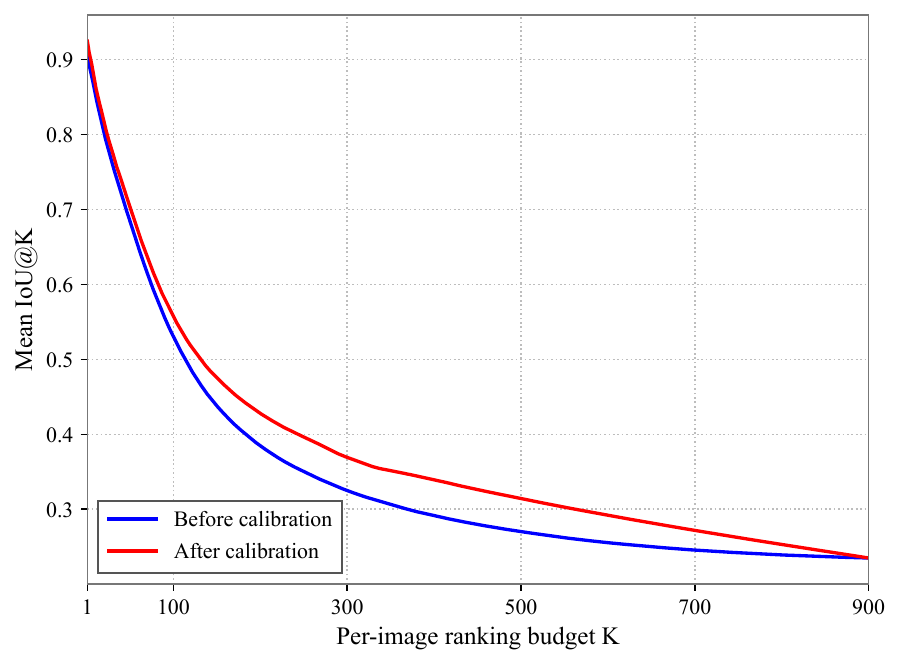}
    \vspace{-0.5cm}
    \caption{Mean class-agnostic maximum IoU of the Top-$K$ queries before and after calibration. $K$ denotes the per-image ranking budget.}
    \label{fig:vis}
\vspace{-0.4cm}
\end{figure}

\subsection{Ablation Studies and Analysis}

\noindent\textbf{Component Effectiveness.}
Table~\ref{tab:module_ablation} shows that CLS-Module and CLC-Module are effective and complementary. CLS-Module improves AP and AP$_{75}$ by 1.0\% and 1.6\%, while CLC-Module improves AP and AP$_{50}$ by 0.9\% and 1.4\%; combining them increases AP and AP$_{75}$ by 1.5\% and 2.0\%, respectively. These gains arise because CLS-Module stabilizes localization quality estimation, while CLC-Module improves classification--localization consistency.

\noindent\textbf{Calibration Strategies.}
As shown in Table~\ref{tab:calibration_strategy}, we compare the adaptive calibration strategy with fixed-exponent fusion, $S_{i,c}^{\mathrm{fix}}=p_{i,c}(Q_i^{\mathrm{geo}})^\eta$, and linear fusion, $S_{i,c}^{\mathrm{lin}}=p_{i,c}[(1-\omega)+\omega Q_i^{\mathrm{geo}}]$. Our method improves AP by 0.4\% over both alternatives. This advantage stems from adaptively adjusting the contribution of geometric quality according to classification reliability, thereby avoiding the limitation of applying a globally fixed coefficient to all candidates.

\noindent\textbf{Hyperparameter Sensitivity.}
Table~\ref{tab:hyperparameter_sensitivity} shows that CLSC-DETR remains stable across the tested parameter ranges. The best performance is obtained with $K=2$, $\alpha=1.0$, $\beta=0.20$, and $\gamma=1.2$. Aggregating excessive candidates or applying overly strong calibration may introduce noise and disturb the classification and localization balance.

\noindent\textbf{Visualization.}
As shown in Fig.~\ref{fig:vis}, the calibrated ranking achieves a higher Mean IoU@$K$ across most ranking budgets, indicating that better-localized queries are ranked higher. These results demonstrate that the proposed calibration strategy improves localization-quality-aware candidate ranking without changing the candidate set.

\section{Conclusion}
\label{sec:con}
This paper proposes CLSC-DETR, a Cross-Layer Local Support and Consistency Calibration framework for UAV small-object detection. By integrating the CLS and CLC modules, CLSC-DETR alleviates localization quality estimation bias and candidate-ranking instability caused by insufficient geometric information. CLS improves small-object localization quality estimation through cross-layer geometric support, while CLC mitigates classification--localization inconsistency through classification-reliability-aware calibration. Compared with MI-DETR, CLSC-DETR improves AP/AP$_{75}$ by 1.5\%/2.0\% on VisDrone and 3.0\%/3.8\% on UAVDT. The current fixed local search and Top-$K$ aggregation may not fully adapt to object-scale variations. Future work will explore learnable neighborhoods and adaptive cross-layer aggregation to further improve representation and generalization.

\vfill\pagebreak

\bibliographystyle{IEEEbib}
\bibliography{strings,refs}

@article{zhu2018visdrone,
  author        = {Zhu, Pengfei and Wen, Longyin and Bian, Xiao and Ling, Haibin and Hu, Qinghua},
  title         = {Vision Meets Drones: A Challenge},
  journal       = {arXiv preprint arXiv:1804.07437},
  year          = {2018},
  eprint        = {1804.07437},
  archiveprefix = {arXiv},
  primaryclass  = {cs.CV}
}

@article{shen2024imagpose,
  title={Imagpose: A unified conditional framework for pose-guided person generation},
  author={Shen, Fei and Tang, Jinhui},
  journal={Advances in neural information processing systems},
  volume={37},
  pages={6246--6266},
  year={2024}
}

@inproceedings{shen2025imagdressing,
  title={Imagdressing-v1: Customizable virtual dressing},
  author={Shen, Fei and Jiang, Xin and He, Xin and Ye, Hu and Wang, Cong and Du, Xiaoyu and Li, Zechao and Tang, Jinhui},
  booktitle={Proceedings of the AAAI Conference on Artificial Intelligence},
  volume={39},
  number={7},
  pages={6795--6804},
  year={2025}
}

@article{zhang2025uavdetr,
  author        = {Zhang, Huaxiang and Liu, Kai and Gan, Zhongxue and Zhu, Guo-Niu},
  title         = {{UAV-DETR}: Efficient End-to-End Object Detection for Unmanned Aerial Vehicle Imagery},
  journal       = {arXiv preprint arXiv:2501.01855},
  year          = {2025},
  eprint        = {2501.01855},
  archiveprefix = {arXiv},
  primaryclass  = {cs.CV}
}

@article{peng2025hedsdetr,
  author        = {Peng, Hongxing and Chen, Lide and Zhu, Hui and Chen, Yan},
  title         = {High-Frequency Semantics and Geometric Priors for End-to-End Detection Transformers in Challenging {UAV} Imagery},
  journal       = {arXiv preprint arXiv:2507.00825},
  year          = {2025},
  eprint        = {2507.00825},
  archiveprefix = {arXiv},
  primaryclass  = {cs.CV}
}

@article{xia2026efsidetr,
  author        = {Xia, Yu and Liu, Chang and Xiang, Tianqi and Tu, Zhigang},
  title         = {{EFSI-DETR}: Efficient Frequency-Semantic Integration for Real-Time Small Object Detection in {UAV} Imagery},
  journal       = {arXiv preprint arXiv:2601.18597},
  year          = {2026},
  eprint        = {2601.18597},
  archiveprefix = {arXiv},
  primaryclass  = {cs.CV}
}

@inproceedings{jiang2018iounet,
  author    = {Jiang, Borui and Luo, Ruixuan and Mao, Jiayuan and Xiao, Tete and Jiang, Yuning},
  title     = {Acquisition of Localization Confidence for Accurate Object Detection},
  booktitle = {European Conference on Computer Vision},
  pages     = {784--799},
  year      = {2018}
}

@inproceedings{carion2020detr,
  author    = {Carion, Nicolas and Massa, Francisco and Synnaeve, Gabriel and Usunier, Nicolas and Kirillov, Alexander and Zagoruyko, Sergey},
  title     = {End-to-End Object Detection with Transformers},
  booktitle = {European Conference on Computer Vision},
  pages     = {213--229},
  year      = {2020}
}

@inproceedings{zhang2023dino,
  author    = {Zhang, Hao and Li, Feng and Liu, Shilong and Zhang, Lei and Su, Hang and Zhu, Jun and Ni, Lionel M. and Shum, Heung-Yeung},
  title     = {{DINO}: {DETR} with Improved DeNoising Anchor Boxes for End-to-End Object Detection},
  booktitle = {International Conference on Learning Representations},
  year      = {2023}
}

@inproceedings{zhao2024rtdetr,
  author    = {Zhao, Yian and Lv, Wenyu and Xu, Shangliang and Wei, Jinman and Wang, Guanzhong and Dang, Qingqing and Liu, Yi and Chen, Jie},
  title     = {{DETR}s Beat {YOLO}s on Real-Time Object Detection},
  booktitle = {Proceedings of the IEEE/CVF Conference on Computer Vision and Pattern Recognition},
  pages     = {16965--16974},
  year      = {2024}
}

@inproceedings{ye2023cascadedetr,
  author    = {Ye, Mingqiao and Ke, Lei and Li, Siyuan and Tai, Yu-Wing and Tang, Chi-Keung and Danelljan, Martin and Yu, Fisher},
  title     = {{Cascade-DETR}: Delving into High-Quality Universal Object Detection},
  booktitle = {Proceedings of the IEEE/CVF International Conference on Computer Vision},
  pages     = {6704--6714},
  year      = {2023}
}

@inproceedings{pu2023rankdetr,
  author    = {Pu, Yifan and Liang, Weicong and Hao, Yiduo and Yuan, Yuhui and Yang, Yukang and Zhang, Chao and Hu, Han and Huang, Gao},
  title     = {{Rank-DETR} for High Quality Object Detection},
  booktitle = {Advances in Neural Information Processing Systems},
  volume    = {36},
  pages     = {39004--39018},
  year      = {2023}
}

@article{cai2023aligndetr,
  author        = {Cai, Zhi and Ravichandran, Avinash and Favaro, Paolo and Wang, Minsi and Modolo, Davide and Bhotika, Rahul and Tu, Zhuowen and Soatto, Stefano},
  title         = {{Align-DETR}: Improving {DETR} with Simple {IoU}-Aware {BCE} and Query Alignment},
  journal       = {arXiv preprint arXiv:2304.07527},
  year          = {2023},
  eprint        = {2304.07527},
  archiveprefix = {arXiv},
  primaryclass  = {cs.CV}
}

@article{munir2023caldetr,
  author        = {Munir, Muhammad Akhtar and Khan, Salman and Khan, Muhammad Haris and Ali, Mohsen and Khan, Fahad Shahbaz},
  title         = {{Cal-DETR}: Calibrated Detection Transformer},
  journal       = {arXiv preprint arXiv:2311.03570},
  year          = {2023},
  eprint        = {2311.03570},
  archiveprefix = {arXiv},
  primaryclass  = {cs.CV}
}

@article{nan2025midetr,
  author        = {Nan, Zhixiong and Li, Xianghong and Dai, Jifeng and Xiang, Tao},
  title         = {{MI-DETR}: An Object Detection Model with Multi-time Inquiries Mechanism},
  journal       = {arXiv preprint arXiv:2503.01463},
  year          = {2025},
  eprint        = {2503.01463},
  archiveprefix = {arXiv},
  primaryclass  = {cs.CV}
}

@inproceedings{he2016resnet,
  author    = {He, Kaiming and Zhang, Xiangyu and Ren, Shaoqing and Sun, Jian},
  title     = {Deep Residual Learning for Image Recognition},
  booktitle = {Proceedings of the IEEE Conference on Computer Vision and Pattern Recognition},
  pages     = {770--778},
  year      = {2016}
}

@inproceedings{loshchilov2019decoupled,
  author    = {Loshchilov, Ilya and Hutter, Frank},
  title     = {Decoupled Weight Decay Regularization},
  booktitle = {International Conference on Learning Representations},
  year      = {2019}
}

@article{wang2024yolov10,
  author  = {Wang, Ao and Chen, Hui and Liu, Lihao and Chen, Kai and Lin, Zijie and Han, Jungong and Ding, Guiguang},
  title   = {{YOLOv10}: Real-Time End-to-End Object Detection},
  journal = {arXiv preprint arXiv:2405.14458},
  year    = {2024}
}

@misc{jocher2024yolo11,
  author       = {Jocher, Glenn and Qiu, Jing},
  title        = {{Ultralytics YOLO11}},
  year         = {2024},
  howpublished = {\url{https://github.com/ultralytics/ultralytics}}
}

@inproceedings{Du2018UAVDT,
  author    = {Du, Dawei and Qi, Yuankai and Yu, Hongyang and Yang, Yifan and Duan, Kaiwen and Li, Guorong and Zhang, Weigang and Huang, Qingming and Tian, Qi},
  title     = {The Unmanned Aerial Vehicle Benchmark: Object Detection and Tracking},
  booktitle = {Proceedings of the European Conference on Computer Vision (ECCV)},
  pages     = {370--386},
  year      = {2018}
}

@article{Zhu2022VisDrone,
  author  = {Zhu, Pengfei and Wen, Longyin and Du, Dawei and Bian, Xiao and Fan, Heng and Hu, Qinghua and Ling, Haibin},
  title   = {Detection and Tracking Meet Drones Challenge},
  journal = {IEEE Transactions on Pattern Analysis and Machine Intelligence},
  year    = {2022},
  volume  = {44},
  number  = {11},
  pages   = {7380--7399},
  doi     = {10.1109/TPAMI.2021.3119563}
}

\end{document}